\documentclass{article}
\usepackage{hyperref}
\usepackage{spconf, amssymb, amsmath, subcaption, ragged2e, booktabs, graphicx, xcolor}

\title{SpeechCLIP+: Self-supervised multi-task representation learning for speech via CLIP and speech-image data}
\name{
\parbox{\linewidth}{
\centering
Hsuan-Fu Wang$^1$, Yi-Jen Shih$^2$, Heng-Jui Chang$^3$, Layne Berry$^2$, Puyuan Peng$^2$, \\
Hung-yi Lee$^1$, Hsin-Min Wang$^4$, and David Harwath$^2$} 
}
\address{
$^1$National Taiwan University, Taiwan \\
$^2$The University of Texas at Austin, USA \\
$^3$Massachusetts Institute of Technology, USA \\
$^4$Institute of Information Science, Academia Sinica, Taiwan\\
}
  
\begin{document}
\ninept
\maketitle
\begin{abstract}

The recently proposed visually grounded speech model SpeechCLIP is an innovative framework that bridges speech and text through images via CLIP without relying on text transcription. On this basis, this paper introduces two extensions to SpeechCLIP. First, we apply the Continuous Integrate-and-Fire (CIF) module to replace a fixed number of CLS tokens in the cascaded architecture. Second, we propose a new hybrid architecture that merges the cascaded and parallel architectures of SpeechCLIP into a multi-task learning framework. Our experimental evaluation is performed on the Flickr8k and SpokenCOCO datasets. The results show that in the speech keyword extraction task, the CIF-based cascaded SpeechCLIP model outperforms the previous cascaded SpeechCLIP model using a fixed number of CLS tokens. Furthermore, through our hybrid architecture, cascaded task learning boosts the performance of the parallel branch in image-speech retrieval tasks.


\end{abstract}

\section{Introduction}
 Recent research in speech processing has focused on self-supervised learning (SSL), which pre-trains upstream models on different pretext tasks~\cite{mohamed2022self}, such as generation or reconstruction~\cite{van2017neural, chung2019apc}, contrastive learning~\cite{baevski2019vq, baevski2020wav2vec}, prediction~\cite{hsu2021hubert, chen2022wavlm}, and knowledge distillation~\cite{chang2022distilhubert, baevski2022data2vec}.
These pre-trained models have been shown to outperform supervised models on certain downstream tasks, like speech recognition~\cite{hsu2021hubert}. In addition to unimodal SSL methods, it may be beneficial to leverage different modalities, e.g., contextual image information can help speech models recognize similar-sounding words, thereby boosting the model's performance on speech recognition~\cite{sun2016look}. 
Therefore, some studies have utilized image-speech or image-text pairs for multi-modal training. 



Speech processing models trained using image-speech paired data are known as Visually Grounded Speech (VGS) models. Many studies~\cite{peng2022word, peng2022fastvgsplus} have found that these VGS models are beneficial for many tasks. The recently proposed SpeechCLIP~\cite{shih2022speechclip} is a VGS model that leverages the pre-trained CLIP~\cite{radford2021learning} image-text model and the speech SSL model HuBERT~\cite{hsu2021hubert}. HuBERT is trained with a masked language modeling strategy, which provides good initialization for general speech processing tasks~\cite{yang2021superb}. CLIP employs contrastive learning to pre-train robust image and text encoders by aligning semantically related images and text captions. 
By aligning speech-image pairs during training, SpeechCLIP learns to transform speech representations into the same embedding space as the pre-trained CLIP, thereby aligning speech and text together without transcriptional supervision. Moreover, M-SpeechCLIP~\cite{berry2022m}, an extension of SpeechCLIP, demonstrates the capability of multilingual speech-to-image retrieval.

\begin{figure*}
     \begin{subfigure}[b]{0.49 \textwidth}
        \centering
        \includegraphics[width=0.75\textwidth,trim={0cm 0cm 0cm 1.3cm}]
    {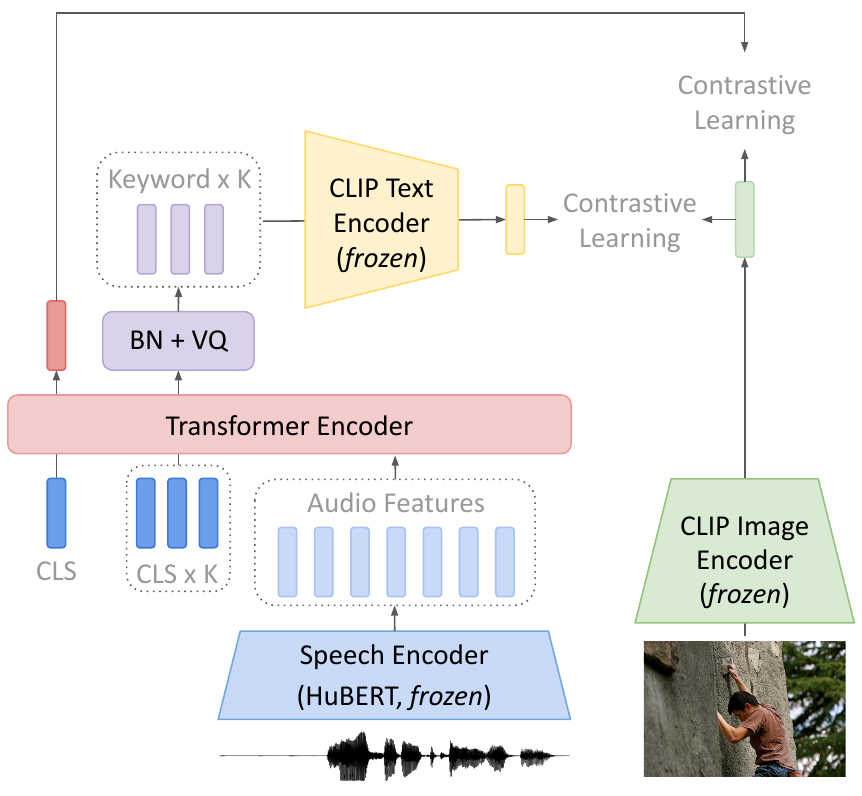}
        \caption{hybrid SpeechCLIP}
        \label{fig:hybrid_arch}
        \medskip
    \end{subfigure}
    \begin{subfigure}[b]{0.49\textwidth}
        \centering
        \includegraphics[width=0.8\textwidth,trim={0cm 0cm 0cm 1.3cm}]
    {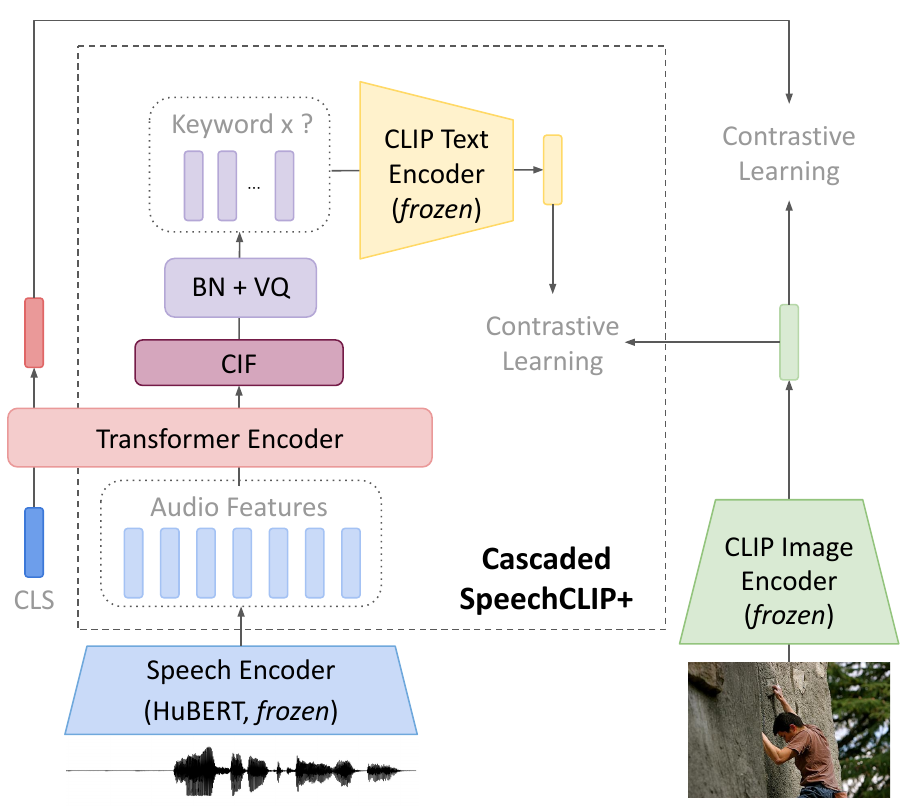}
        \caption{hybrid SpeechCLIP+ with cascaded SpeechCLIP+ }
        \label{fig:hybrid+_arch}
        \medskip
    \end{subfigure}
    \vspace{-0.3cm}
    \caption{Illustration of the proposed models. BN and VQ denote batch normalization and vector quantization processes, respectively. (a) In hybrid SpeechCLIP, the training loss combines the contrastive loss between the leftmost CLS token and the output image representation of the CLIP image encoder (the parallel branch same as parallel SpeechCLIP~\cite{shih2022speechclip}) and the contrastive loss between the output speech representation of the CLIP text encoder for the remaining $K$ CLS tokens and the output image representation of the CLIP image encoder (the cascaded branch same as cascaded SpeechCLIP~\cite{shih2022speechclip}). (b) In cascaded SpeechCLIP+, instead of extracting keyword information through a fixed number of learnable CLS tokens, CIF is used to segment frame-level features into subword-level keyword sequences. In hybrid SpeechCLIP+, the parallel branch is based on parallel SpeechCLIP, and the cascaded branch is based on cascaded SpeechCLIP+.}
    \vspace{-0.3cm}
    \label{fig:model_comparison}
\end{figure*}

SpeechCLIP employs two architectures: parallel and cascaded.
Parallel SpeechCLIP aligns semantically related images and spoken captions with utterance-level information, while cascaded SpeechCLIP aligns them with intermediately extracted subword-level information. Each architecture has its strengths: cascaded SpeechCLIP is capable of extracting a fixed number of keywords directly from speech without any text supervision and image tagging system. On the other hand, parallel SpeechCLIP excels at image-speech retrieval tasks. In this paper, we aim to further improve two types of architectures in terms of their strengths. Regarding the keyword extraction task, we have identified some issues in~\cite{shih2022speechclip}. First, the $K$ CLS tokens in the original cascaded architecture tend to attend to similar segments in the input speech, resulting in duplicate keywords in the output. Additionally, the fixed number of CLS tokens may not be flexible enough when the duration of the input speech varies significantly.

To address these issues, we apply the Continuous Integrate-and-Fire (CIF)~\cite{dong2020cif} module to replace a fixed number of CLS tokens in the original cascaded architecture. We believe that CIF's monotonic alignment method will alleviate the issue of duplicate keywords. Furthermore, CIF enables the ability to output a dynamic number of keywords. The experimental results also manifest improvement when we do not consider duplicate keywords in the speech keyword extraction task. For the image-speech retrieval tasks, we posit that meaningful subwords' information could be beneficial in addition to the summary of the entire utterance. Thus, we propose a hybrid architecture, a multi-task learning framework that integrates the learning objectives of SpeechCLIP from both cascaded and parallel architectures. Experimental results manifest the improvements on Flickr8k~\cite{rashtchian2010collecting}.

\vspace{-0.3cm}
\section{Method}

As with the training of SpeechCLIP~\cite{shih2022speechclip}, the input to the model is a batch of paired images and audio waveforms. Images are passed to CLIP's image encoder to extract image features, and audio waveforms are passed to a pre-trained SSL speech encoder. CLIP's text and image encoders are frozen during SpeechCLIP training. They serve as projectors of the pre-trained image-text semantic space. We follow SpeechCLIP by using HuBERT as our speech encoder. We also freeze it during model training but employ a set of learnable weights to perform a weighted sum of all its hidden states for audio feature extraction. 


\subsection{Continuous Integrate-and-Fire (CIF)} \label{cif}
CIF is a soft, monotonic alignment method used to segment input sequential features with assigned weights. Its effectiveness has been demonstrated in previous speech segmentation tasks~\cite{meng2022compressing, fan2022sequence}. Given sequential features $\mathbf{x} = (x_1, x_2, \cdots, x_T)$, CIF uses a convolution layer followed by a feed-forward layer with a sigmoid function to generate the corresponding weights $\boldsymbol{\alpha} = (\alpha_1, \alpha_2, \cdots, \alpha_T)$. Segmentation boundaries are determined by accumulating $\alpha_t$. As a result, CIF generates a sequence $\mathbf{c} = (c_1, c_2, \cdots, c_{L})$ based on its segmentation and aggregation process. 

During training, for each input speech, a target segmentation number $L$ is required, which determines the desired output length of $\mathbf{c}$. Therefore, in~\cite{dong2020cif}, the following quantity loss $\mathcal{L}_{\text{QUA}}$ is used to encourage CIF to produce the correct cumulative sum of $\boldsymbol{\alpha}$,
\begin{equation}
    \label{eq:q_loss}
    \mathcal{L}_{\text{QUA}} = \left| \sum_{t=1}^{T} \alpha_t - L \right|.
\end{equation}
Additionally, since the segmentation number may sabotage the training stability, in~\cite{dong2020cif},
 a scaling strategy is used to generate $\boldsymbol{\alpha'} = (\alpha'_1, \alpha'_2, \cdots, \alpha'_T)$, where 
 \begin{equation}
    \label{eq:scale}
     \alpha'_j = \frac{\alpha_j}{\sum_{t=1}^{T} \alpha_{t}} \times L,
 \end{equation}
 for performing segmentation and aggregation. The scaling strategy ensures that the output length of $\mathbf{c}$ is adjusted to the desired $L$.

\subsection{Cascaded SpeechCLIP+}\label{c+_sec}

As shown in Fig.~\ref{fig:hybrid+_arch}, in cascaded SpeechCLIP+, instead of extracting keyword information through a fixed number of learnable CLS tokens, we apply CIF mechanism to segment frame-level features into a subword-level keyword sequence. During training, we set the target length of the quantity loss in Eq. (\ref{eq:q_loss}) to 5\% of the input length, based on the average ratio between the length of the Byte-Pair Encoding (BPE) token sequence and the output feature length of HuBERT, as observed during our experimentation. Additionally, we apply the scaling strategy in Eq. (\ref{eq:scale}) to enhance training stability in the first 5k steps.
After the CIF mechanism, the same as cascaded SpeechCLIP, we employ batch normalization and vector quantization to generate BPE tokens. These tokens are then input into the CLIP text encoder to extract representations, which are used to generate representations for computing contrastive loss $\mathcal{L}_{\text{cascaded}}$ with image features. The overall loss $\mathcal{L}$ for training cascaded SpeechCLIP+ is the linear combination of $\mathcal{L}_{\text{cascaded}}$ and $\mathcal{L}_{\text{QUA}}$,
\begin{equation}
    \label{eq:c+_loss}
    \mathcal{L} = \lambda_{c} \times \mathcal{L}_{\text{cascaded}} + \lambda_{q} \times \mathcal{L}_{\text{QUA}},
\end{equation}
where $\lambda_{c}$ and $\lambda_{q}$ are adjustable weights.

\subsection{Hybrid SpeechCLIP}\label{hybrid_sec}
Two types of SpeechCLIP are expected to benefit speech encoders from semantically related images through different alignment methods. Parallel SpeechCLIP enables speech encoders to benefit from utterance summarization, while cascaded SpeechCLIP enables speech encoders to benefit from capturing subword-level information from utterances.  The advantages of combining the two approaches are quite intuitive. As shown in Fig.~\ref{fig:hybrid_arch}, there is a total $K+1$ CLS tokens in hybrid SpeechCLIP, the parallel branch refers to the path with the leftmost single CLS, while the cascaded branch refers to the path with the remaining $K$ CLS tokens. The parallel branch is the same as the parallel SpeechCLIP, the first CLS representation of the transformer encoder's output is used to compute the contrastive loss $\mathcal{L}_{\text{parallel}}$ with image features. The cascaded branch is also the same as the cascaded SpeechCLIP, the remaining $K$ CLS representations of the transformer encoder's output will be batch-normalized to match the mean and variance of CLIP's BPE token embeddings and vector-quantized into BPE tokens and then used as input for the CLIP text encoder. The output of the text encoder is used to compute the contrastive loss $\mathcal{L}_{\text{cascaded}}$ with image features. The parallel and cascaded branches are trained jointly, and the overall loss $\mathcal{L}$ is the linear combination of $\mathcal{L}_{\text{parallel}}$ and $\mathcal{L}_{\text{cascaded}}$,
\begin{equation}
    \label{eq:h_loss}
    \mathcal{L} = \lambda_{p} \times \mathcal{L}_{\text{parallel}} + \lambda_{c} \times \mathcal{L}_{\text{cascaded}},
\end{equation}
where $\lambda_{p}$ and $\lambda_{c}$ are adjustable weights.

\subsection{Hybrid SpeechCLIP+}
As shown in Fig.~\ref{fig:hybrid+_arch}, in hybrid SpeechCLIP+, the parallel branch is the parallel SpeechCLIP model, and the cascaded branch is the cascaded SpeechCLIP+ model. In the parallel branch, we apply one CLS token to compute the contrastive loss $\mathcal{L}_{\text{parallel}}$ with image features. In the cascaded branch, same as cascaded SpeechCLIP+, we apply CIF to segment frame-level features into a subword-level keyword sequence. During training, we set the target length of the quantity loss in Eq. (\ref{eq:q_loss}) to 5\% of the input length and apply the scaling strategy in Eq. (\ref{eq:scale}) in the first 5k steps. After the CIF mechanism, we employ batch normalization and vector quantization to generate BPE tokens. These tokens are then input into the CLIP text encoder to extract representations, which are subsequently used to generate representations for computing the contrastive loss $\mathcal{L}_{\text{cascaded}}$ with image features. The parallel and cascaded 
branches are trained jointly with the overall loss $\mathcal{L}$ as follows,
\begin{equation}
    \label{eq:h+_loss}
    \mathcal{L} = \lambda_{p} \times \mathcal{L}_{\text{parallel}} + \lambda_{c} \times \mathcal{L}_{\text{cascaded}} + \lambda_{q} \times \mathcal{L}_{\text{QUA}},
\end{equation}
where $\lambda_{p}$, $\lambda_{c}$, and $\lambda_{q}$ are adjustable weights.

\begin{figure*}
    \centering
    \includegraphics[width=0.9\textwidth, trim={0cm 0cm 0cm 0cm}]{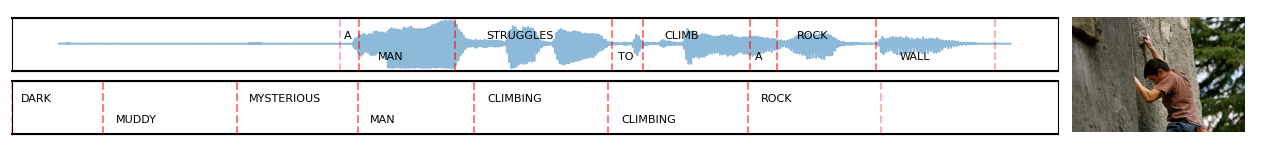}
    \vspace{-0.1cm}
    \caption{An example of keywords extracted by Cascaded SpeechCLIP+ from the Flickr8k test set, showing the image, spoken caption, and extracted keywords with corresponding segments.}
    \label{fig:keywords}
    \vspace*{-10pt}
\end{figure*}

\section{Experiments and Results}
\subsection{Experimental setups}
\textbf{Datasets.} 
Our models are trained on two image-audio datasets, namely the Flickr8k Audio Captions Corpus~\cite{rashtchian2010collecting} and SpokenCOCO~\cite{karpathy2015deep}. 
Each image in both datasets is paired with five spoken captions collected by humans reciting the corresponding text captions. 
The training, development, and test sets in Flickr8k contain 30k, 5k, and 5k utterances, respectively, while SpokenCOCO has 565k, 25k, and 25k utterances in its training, development, and test sets. Following SpeechCLIP~\cite{shih2022speechclip}, we use the Karpathy split for SpokenCOCO~\cite{karpathy2015deep}.

\begin{table}[th]
  \small
  \caption{BPE extraction performance on the Flickr8k and SpokenCOCO test sets.}
  \label{tab:BPE-Flickr8k}
  \vspace*{-5pt}
  \centering
  \begin{tabular}{ l@{~~}c@{~~}c@{~~}c@{~~}c@{~}c@{~~}c@{~~}c@{~~} }
    \toprule
    & \multicolumn{3}{c}{w/o stop words} & & \multicolumn{3}{c}{w/ stop words}\\
    \cmidrule{2-4}
    \cmidrule{6-8}
    Model& R & P & F1 & & R & P & F1\\
    \midrule
    & \multicolumn{7}{c}{Flickr8k} \\
    \cmidrule{2-8}
    Cascaded~\cite{shih2022speechclip}&
    7.39 & 0.94 & 1.66 & & 8.3  & 1.68 & 2.79\\
    Cascaded+& \textbf{27.16} & \textbf{3.55} & \textbf{6.29} & & \textbf{18.11} & \textbf{3.62} & \textbf{6.04}\\
    Cascaded(h) & 7.01 & 0.85 & 1.52 & & 5.41 & 1.06 & 1.77\\
    Cascaded(h)+ & 16.52 & 2.16 & 3.82 & & 11.40 & 2.23 & 3.73\\
    \midrule
    Cascaded Large~\cite{shih2022speechclip}&
    5.27 & 0.75 & 1.31 & & 14.84 & 3.10 & 5.13\\
    Cascaded+ Large& \textbf{27.27} & \textbf{3.73} & \textbf{6.56} & & \textbf{20.74} & \textbf{4.2} & \textbf{6.99}\\
    Cascaded(h) Large & 6.73 & 0.86 & 1.53 & & 6.18 & 1.25 & 2.08\\
    Cascaded(h)+ Large & 21.69 & 2.92 & 5.15 & & 15.48 & 3.08 & 5.14\\
    \midrule
        & \multicolumn{7}{c}{SpokenCOCO} \\
    \cmidrule{2-8}
    Cascaded Large~\cite{shih2022speechclip}&
    2.08 & 0.39 & 0.65 & & 12.96  & 2.89 & 4.72\\
    Cascaded+ Large& 20.30 & 2.74 & 4.83 & & 18.42 & 3.39 & 5.73\\
    Cascaded(h) Large& 1.68 & 0.45 & 0.71 & & 20.37 & 4.48 & 7.35\\
    Cascaded(h)+ Large& \textbf{20.56} & \textbf{3.73} & \textbf{6.32} & & \textbf{26.21} & \textbf{4.95} & \textbf{8.33} \\
    \bottomrule
  \end{tabular}
  \vspace{-0.3cm}
\end{table}

\noindent\textbf{Compared models.} In the following experiments, the baseline cascaded SpeechCLIP, the proposed cascaded SpeechCLIP+, cascaded SpeechCLIP in hybrid SpeechCLIP, and cascaded SpeechCLIP+ in hybrid SpeechCLIP+ are denoted as Cascaded, Cascaded+, Cascaded(h) and Cascaded(h)+, respectively.
Parallel SpeechCLIP can be trained jointly with cascaded SpeechCLIP in hybrid SpeechCLIP or cascaded SpeechCLIP+ in hybrid SpeechCLIP+. The model trained in the former way is represented as Parallel(h) to distinguish it from the Parallel(h)+ model trained in the latter way. Both models are
compared with the baseline parallel SpeechCLIP (denoted as Parallel). All models are trained using the base or large model size setting in~\cite{shih2022speechclip}. When the large model size setting is used, the model is marked ``Large'', e.g., Cascaded+ Large. 

\noindent\textbf{Implementation details.} We use the same transformer encoder and the CLIP~\cite{radford2021learning} model as in SpeechCLIP~\cite{shih2022speechclip}. The convolution layer of CIF consists of a single one-dimensional convolution with $d_{model}$ channels, a stride of 1, and a kernel width of 3. Here, $d_{model}$ is set to 768 for the base models and 1024 for the large models. The one-dimensional convolution is followed by a dropout with a probability of 0.5 and a ReLU activation function. Regarding the loss weights in Eqs. (\ref{eq:c+_loss}), (\ref{eq:h_loss}), and (\ref{eq:h+_loss}), we set $\lambda_c = \lambda_p = 1.0$ and $\lambda_{q} = 0.25$. We employ the same optimization strategy as in SpeechCLIP~\cite{shih2022speechclip}. All models are trained with a batch size of 256 using two NVIDIA RTX 3090 GPUs with 24GB of memory. The hybrid+ Large model trained on SpokenCOCO took 4 days to converge, while model training under other conditions took less than two days.


\subsection{Keyword extraction}
As discussed in SpeechCLIP~\cite{shih2022speechclip}, we evaluate the keyword extraction ability of cascaded SpeechCLIP+ through qualitative and quantitative analyses. Although there are other some previous works~\cite{kamper2017visually, Nortje2023visuallyprompt} on unsupervised Bag of Words prediction, it is challenging to compare them in this task because we use different levels of tokens (BPE tokens) for training, whereas they use word tokens. Moreover, they require an image tagging system, while we do not. So we only compare with~\cite{shih2022speechclip} in this section.

For qualitative analysis, from the example from the Flickr8k test set in Fig. 2, we can observe the ability of cascaded SpeechCLIP+ to capture semantically aligned keywords and their corresponding boundaries in the input speech. For example, our model successfully extracted
the words “MAN” and “ROCK” and found their fragments in the
audio waveform. In addition, the extracted word “CLIMBING” is
semantically correlated to "CLIMB".


\begin{table}[th]
  \small
  \caption{Word and BPE extraction performance of Cascaded+ Large on the Flickr8k test set. ``Type'' refers to the granularity of the units considered, while ``Top $K$'' represents the top $K$ BPEs retrieved by each CLS.}
  \label{tab:word-Flickr8k}
  \vspace*{-5pt}
  \centering
  \begin{tabular}{ l@{~~}c@{~~}c@{~~}c@{~~}c@{~~}c@{~}c@{~~}c@{~~}c@{~~} }
    \toprule
    & & \multicolumn{3}{c}{w/o stop words} & & \multicolumn{3}{c}{w/ stop words}\\
    \cmidrule{3-5}
    \cmidrule{7-9}
    Type&Top $K$& R & P & F1 & & R & P & F1\\
    \midrule
    Word & 1 & 19.81 & \textbf{12.06} & \textbf{14.99} & & 14.12 & \textbf{13.66} & \textbf{13.89}\\
         & 2 & 22.80 & 6.84 & 10.53 & & 16.54 & 7.86 & 10.65\\
         & 3 & 25.07 & 4.95 & 8.27 & & 18.74 & 5.85 & 8.91 \\
         & 4 & 26.76 & 3.92 & 6.83 & & 19.97 & 4.61 & 7.49 \\
         & 5 & \textbf{27.59} & 3.20 & 5.74 & & \textbf{20.86} & 3.82 & 6.45 \\
    \midrule
    BPE  & 1 & 19.17 & \textbf{12.99} & \textbf{15.49} & & 13.79 & \textbf{13.97} & \textbf{13.88} \\
         & 2 & 22.29 & 7.53 & 11.26 & & 16.29 & 8.25 & 10.96 \\
         & 3 & 24.52 & 5.54 & 9.03 & & 18.46 & 6.24 & 9.32 \\
         & 4 & 26.31 & 4.46 & 7.63 & & 19.77 & 5.01 & 7.99 \\
         & 5 & \textbf{27.27} & 3.73 & 6.56 & & \textbf{20.74} & 4.2 & 6.99\\
    \bottomrule
  \end{tabular}
\end{table}


For the quantitative analysis, the results are shown in Table \ref{tab:BPE-Flickr8k}. In this experiment, the top 5 BPEs closest in cosine similarity to each CLS token (for Cascaded and Cascaded(h)) or each CIF-segmented token (for Cascaded+ and Cascaded(h)+) are retrieved. Performance is evaluated in terms of recall, precision, and F1-score to the BPE sequence of the transcription of the spoken caption. Considering that stop words in a sentence usually have little impact on the semantics of the sentence, we provide complete evaluation results (see w/ stop words) and evaluation results that ignore stop words (see w/o stop words). Stop words are pronouns, articles, prepositions, and conjunctions. We use the stop word set from the nltk python package.

It is clearly seen from the table that the proposed Cascaded+ models significantly outperform their corresponding baseline Cascaded models (Cascaded+ vs Cascaded, Cascaded(h)+ vs Cascaded(h), Cascaded+ Large vs Cascaded Large, and Cascaded(h)+ Large vs Cascaded(h) Large). 
Surprisingly, joint training of parallel and cascaded branches did not always bring performance improvements to Cascaded and Cascaded+ models. The reason remains to be further studied. 

Table \ref{tab:word-Flickr8k} shows the performance of word extraction. Since the previous experiment shows that cascade SpeechCLIP+ Large (Cascaded+ Large) performs best on the Flickr8k test set, we use it in this experiment. We use the extracted neighboring BPEs to construct words. Top $K$ BPEs with $K$ from 1 to 5 are evaluated. The corresponding BPE extraction performance is also provided for reference. As can been seen from Table \ref{tab:word-Flickr8k}, Top 1 BPE gives the best F1 score for word extraction with or without considering stop words. An increase in $K$ can slightly improve the recall rate (R), but decrease the precision rate (P). BPE extraction performance has the same trend as word extraction performance.

\begin{table}[th]
  \small
  \caption{Image-speech retrieval performance on the Flickr8k and SpokenCOCO test sets. Parallel(h) refers to the parallel branch in hybrid SpeechCLIP (Fig.~\ref{fig:hybrid_arch}), while Parallel(h)+ refers to the parallel branch in hybrid SpeechCLIP+ (Fig.~\ref{fig:hybrid+_arch}).}
  \label{tab:image-speech IR}
  \vspace*{-5pt}
  \centering
  \begin{tabular}{ l@{~~}c@{~~}c@{~~}c@{~~}c@{~}c@{~~}c@{~~}c@{~~} }
    \toprule
    & \multicolumn{3}{c}{Speech $\rightarrow$ Image} & & \multicolumn{3}{c}{Image $\rightarrow$ Speech}\\
    \cmidrule{2-4}
    \cmidrule{6-8}
    Model& R@1 & R@5 & R@10 & & R@1 & R@5 & R@10\\
    \midrule
        & \multicolumn{7}{c}{Flickr8k} \\
    \cmidrule{2-8}
    Parallel~\cite{shih2022speechclip}&
    26.7 & 57.1 & 70.0 & & \textbf{41.3} & \textbf{73.9} & \textbf{84.3}\\
    Parallel(h) & \textbf{29.8} & \textbf{60.8} & 73.5 & & 40.9 & 71.6 & 83.9\\
    Parallel(h)+ & 29.3 & 60.1 & \textbf{73.7} & & 39.7 & 72.8 & 83.4\\
    \midrule
    Parallel Large~\cite{shih2022speechclip}&
    39.1 & 72.0 & 83.0 & & \textbf{54.5} & 84.5 & 93.2\\
    Parallel(h) Large& \textbf{43.1} & \textbf{75.6} & \textbf{85.2} & & 54.3 & 85.1 & 93.5\\
    Parallel(h)+ Large& 41.7 & 73.7 & 84.1 & & 54.2 & \textbf{86.8} & \textbf{94.2}\\
        \midrule
        & \multicolumn{7}{c}{SpokenCOCO} \\
    \cmidrule{2-8}
        FaST-VGS$_{\text{CTF}}$~\cite{peng2022fast} & 35.9 &  66.3 &  77.9 & & 48.8 &  78.2 & 87.0 \\
        Parallel Large~\cite{shih2022speechclip} & 35.8 & \textbf{66.5} & \textbf{78.0} & & 50.6 & \textbf{80.9} & \textbf{89.1}\\
    \midrule
    Parallel(h) Large& 32.5 & 60.9 & 72.9 & & 44.2 & 73.9 & 83.8\\
    Parallel(h)+ Large& \textbf{36.5} & 66.3 & 77.9 & & \textbf{51.0} & 80.0 & 88.5\\
    \bottomrule
  \end{tabular}
\end{table}

\subsection{Image-Speech retrieval} 
Next, we evaluate Parallel(h)+ and Parallel(h) on image-speech retrieval tasks. The ``Speech $\rightarrow$ Image'' task is to retrieve the corresponding image given a spoken caption, and the ``Image $\rightarrow$ Speech'' task is to retrieve the corresponding spoken caption given an image. 

From Table \ref{tab:image-speech IR}, we can see that on the Frickr8k dataset, the parallel branches in hybrid SpeechCLIP are mostly better than the corresponding baseline parallel models (Parallel(h) vs Parallel, Parallel(h)+ vs Parallel, Parallel(h) Large vs Parallel Large, and Parallel(h)+ Large vs Parallel Large). The results show that through a hybrid architecture, cascaded task learning improves the performance of parallel branches in image-speech retrieval tasks. The performance of Parallel(h) and Parallel(h)+ is comparable, suggesting that both cascaded SpeechCLIP and cascaded SpeechCLIP+ can effectively enhance the parallel branch in joint training. 
However, on the SpokenCOCO dataset, the performance of Parallel(h) Large and Parallel(h)+ Large is disappointing, especially Parallel(h) Large. The results in Table \ref{tab:BPE-Flickr8k} show that Cascaded(h) Large and Cascaded(h)+ Large trained on SpokenCOCO have a relatively better ability to extract stop word information than models under other experimental settings. This may make Parallel(h) Large (or Parallel(h)+ Large) dominated by stop words captured by Cascaded(h) Large (or Cascaded(h)+ Large) jointly trained under the hybrid architecture.


\section{Conclusions}
In this paper, we attempt to boost the performance of pre-trained speech models on downstream tasks by leveraging visual content and other pre-trained modalities. We propose two extensions to SpeechCLIP. First, we apply the Continuous Integrate-and-Fire (CIF) module to replace a fixed number of CLS tokens in the cascaded architecture. Second, we propose a new hybrid architecture that merges the cascaded and parallel architectures of SpeechCLIP into a multi-task learning framework. In the keyword extraction task, our cascaded SpeechCLIP+ model significantly outperforms the baseline cascaded SpeechCLIP model~\cite{shih2022speechclip}. Experimental results show that using CIF-segmented representations are more effective and flexible than adding a fixed number of CLS tokens when extracting subword and word information in speech. In the image-speech retrieval task, experimental results show that both cascaded SpeechCLIP and cascaded SpeechCLIP+ can effectively enhance the parallel branch in hybrid SpeechCLIP through joint training. In future work, we will investigate other unsupervised speech segmentation~\cite{meng2022compressing} and multi-task learning methods~\cite{li2022blip}. 


\vfill\pagebreak
\bibliographystyle{IEEEbib}
\bibliography{ref}

\end{document}